\documentclass{llncs}

\usepackage[english]{babel}
\usepackage[utf8x]{inputenc}
\usepackage[T1]{fontenc}
\usepackage{floatrow}
\usepackage{siunitx}
\usepackage{adjustbox}
\usepackage{subcaption}
\usepackage{blindtext}
\usepackage{multirow}
\usepackage{tabularx}
\usepackage{array}
\usepackage[a4paper,top=3cm,bottom=2cm,left=3cm,right=3cm,marginparwidth=1.75cm]{geometry}

\usepackage{amsmath}
\usepackage{graphicx}
\usepackage{wrapfig}
\usepackage[colorinlistoftodos]{todonotes}
\usepackage[colorlinks=true, allcolors=blue]{hyperref}
\usepackage{xcolor}
\usepackage{xspace}
\usepackage{color}
\usepackage{algpseudocode}
\usepackage[linesnumbered,ruled]{algorithm2e}
\graphicspath{ {images/} }
\title{Real-time pedestrian recognition on low computational resources}

\author{Guifan Weng$^{1*}$}

\institute{
$^1$Viterbi School of Engineering, University of Southern California, USA\\
}

\begin{document}
\maketitle

\begin{abstract}

Pedestrian recognition has successfully been applied to security, autonomous cars, Aerial photographs. For most applications, pedestrian recognition on small mobile devices is important. However, the limitations of the computing hardware make this a challenging task. In this work, we investigate real-time pedestrian recognition on small physical-size computers with low computational resources for faster speed. This paper presents three methods that work on the small physical size CPUs system. First, we improved the Local Binary Pattern (LBP) features and Adaboost classifier. Second, we optimized the Histogram of Oriented Gradients (HOG) and Support Vector Machine. Third, We implemented fast Convolutional Neural Networks (CNNs). The results demonstrate that the three methods achieved real-time pedestrian recognition at an accuracy of more than 95\% and a speed of more than 5 fps on a small physical size computational platform with a 1.8 GHz Intel i5 CPU. Our methods can be easily applied to small mobile devices with high compatibility and generality.

\end{abstract}

\section{Introduction}
\label{sec:int}

\par Computer vision has been used widely in a variety of applications including medical, military, industry, services, and scientific research \cite{lan2022vision}. In particular, pedestrian recognition in images and videos is a challenging task that attracts the attention of the scientific community and industry alike. It is important in a wide range of applications that intersect with many aspects of our lives: surveillance systems and airport security, autonomous driving and driver assistance systems in high-end cars \cite{zhu2022crowdsensing,xu2019online}, human-robot interaction and immersive, interactive entertainments, smart homes and assistance for senior citizens that live alone, and people-finding for military applications \cite{wu2013c4}. It is also a prerequisite for tasks on mobile devices. For most mobile devices, however, there are some specific constraints that make pedestrian recognition particularly challenging, i.e., limited computational resources, limited physical size, and limited energy. For example, the drones proposed in \cite{lan2016ActionUAV} have limited space inside to fit a small battery and a small computer with low computational resources. 
The same is true for other types of mobile devices like drones \cite{lan2016development,xiang2016uav}, and modular robots \cite{lan2018directed,lan2021learning}.
These constraints make real-time pedestrian recognition on such small physical-size mobile devices a problematic challenge. 

\par In this paper we focus on pedestrian recognition, i.e., people assuming poses that are common while standing or walking. Pedestrian recognition is complex mostly because of the high variability that characterizes the pedestrians' projections on the camera image plane. The appearance of a pedestrian on the image is influenced by the person’s pose, his or her clothing, occlusions, and the atmospheric conditions that contribute to the illumination of the scene \cite{Taiana2013AnImproved}. Background clutter also plays a role in making the detection difficult. That is, the diversity of pedestrians in nature makes real-time pedestrian recognition on low computational hardware a difficult challenge.

\par Although Convolutional Neural Networks (CNNs) became the state-of-the-art approaches with high accuracy for object recognition \cite{sun2023marine,sun2022multi} including pedestrian recognition \cite{He2016Deep}, CNNs take expensive computation. A variety of studies were proposed for real-time pedestrian recognition with faster speed \cite{xiang2016uav,lan2016development,lan2018real}, however, all of them work on the GPU system that outperforms the regular CPU platform. Therefore, we propose and compare three methods that work on the CPU system with small physical size and low computation resources for real-time pedestrian recognition. 

\par After reviewing and considering the state-of-the-art methods in real-time pedestrian recognition, this paper presents three approaches including improved LBP features and AdaBoost classifier, improved HOG features and SVM classifier, and optimized convolutional neural network \cite{lan2019evolving}. First, we use an improved method based on the LBP features and AdaBoost classifier because the LBP features perform well to extract the contour features of pedestrians and the fast speed of the AdaBoost classifier. Second, this paper presents the improved HOG features and SVM classifier. We optimized the speed of HOG features, in particular, the improved approach outperform the original method \cite{Dalal2005Histograms} with the exhaustive search. Third, We achieve a real-time CNN  for pedestrian recognition by optimizing the hyper-parameters. 

\par To assess how our methods work for real-time pedestrian recognition, we investigate the accuracy and speed under the following conditions:
\par 1) The speed must be above 5 fps to meet the requirement that it is real-time.
\par 2) The accuracy must be higher than $95\%$.

\par Therefore, the main contribution in this paper presents three methods for real-time pedestrian recognition on small mobile devices with low computational and small physical size hardware. The remainder of this paper is structured as follows. In section II, we describe related work in object recognition, particularly in real-time pedestrian recognition. A detailed description of our methods is described in section III. We present the experimental results and analyze and discuss these results in section IV. Last, we conclude this paper and provide an outlook on future research.

\section{Related work}
\label{sec:rel}

\par As one of the early real-time object recognition techniques P. Viola and M. Jones \cite{Viola2001Rapid,Viola2004robust} proposed the famous method of simple features and cascade AdaBoost classifier for rapid object detection. Although it worked fast on a desktop computer with Intel Pentium III, it was proposed for face detection and did not work well for pedestrian recognition. In 2005, N. Dalal and B. Triggs \cite{Dalal2005Histograms} proposed the classic solution of Histograms of Oriented Gradients (HOG) features and a Support Vector Machine (SVM) classifier for human detection. This method became popular in object recognition for human detection. The HOG features and linear SVM classifiers are fast and accurate. However, the exhaustive search component is computationally expensive so it is not fast enough on low-performance computational hardware. Therefore, we propose an optimization for faster speed on a low-performance system.

\par In recent years, CNNs became the state-of-the-art methods for object recognition \cite{He2016Deep}, R-CNN\cite{Girskick2014Rich}, Fast R-CNN\cite{Girshick2015fastRCNN}, Faster R-CNN\cite{Ren2017FasterRCNN}, YOLO\cite{Redmon2016YOLO}, SSD\cite{Liu2016SSD}. Although these methods have high accuracy, they usually take expensive computation and work on a GPU system. They are far from working on low-performance computational hardware \cite{lan2021learning2,lan2022time}.

\par Both early and recent popular methods have been widely and successfully applied in many areas, such as pedestrian recognition, face recognition, etc. However, there is a common issue, all of them require expensive computation, hence they only work on powerful computational hardware like GPU systems. A real-time solution on the DSP-based embedded system was proposed in \cite{Arth2008Real-time}. This solution works only on the dedicated DSP hardware platform, not on a general CPU system. An exciting solution was proposed based on HOG features and an SVM classifier for pedestrian detection at 135 fps on a desktop computer equipped with an Intel Core i7 870 and a GPU \cite{VanGool2012Pedestrian}. Similarly, \cite{Nawaz2015HOG-SVM} presented an implementation of vehicle recognition at 4 fps at a resolution of 1224 by 370 pixels based on the HOG feature and linear SVM classifier. However, none of these methods is fast enough on low-performance systems.

An interesting convolutional neural network for object recognition was proposed in \cite{Mao2017Towards}. Although this work optimized fast R-CNN on an embedded platform for real-time object recognition, it works at 1.85fps speed on the CPU and GPU system. Similarly, a low-complexity fully convolutional neural network that works on a weak GPU platform was proposed in \cite{Tripathi2017LCDet} for object recognition based on YOLO. Yet, none of \cite{Tripathi2017LCDet} and \cite{Mao2017Towards} is fast enough on a small-size CPU system. 
\par An alternative approach was proposed in \cite{lan2019evolutionary,lan2019simulated} that implemented real-time object recognition using wireless communication between the mobile robot and the servers. \cite{Lee2017realtime} implemented the near real-time object recognition for drones by offloading the computation onto an off-board computation cloud. Although using a server with powerful computational resources can be a feasible solution, several applications need methods that can operate independently without a server. For instance, the communication time between the drones and servers is affected by variations in wireless bandwidths and this can form a severe bottleneck \cite{Nimmagadda2010Real-time}.

\par Although some of the above methods have fast speed and high accuracy, they still do not work well given the constraints related to small-size CPU computation systems. In addition, there are many studies that work on different low computational hardware for real-time object recognition. An end-to-end deep vision network model was proposed to predict possible good grasps, which works in real-time on a Baxter robot at a rate of 80 frames per second using a GPU system \cite{Guo2017Deepvision}. 
\cite{huang2018Visual} used the exclusive Qualcomm Snapdragon Flight board embedded in a 2.4GHz processor to implement a visual-inertial drone system for real-time moving object detection. Although this computational board has good performance and small size, it is a propriety system exclusive for the drone and the method only recognizes moving objects.

\par In summary, existing methods usually rely on powerful computing resources. In spite of this, some of the algorithms using convolutional neural networks and linear SVM classifiers are reasonably fast, yet not fast enough to implement real-time pedestrian recognition on a regular CPU system with low computation resources.

\section{Methodology}
\label{sec:Meth}
\par This work aims to implement Real-time pedestrian recognition on low computational resources. This paper presents three approaches to achieve the requirements of the speed at 2 fps and the accuracy at 95\%. LBP is a fast algorithm to extract the contour features of the objects, in particular, it is very suitable for extracting the features of pedestrians. The Adaboost classifier is a fast classifier. The method of HOG features and SVM classifier is popular in pedestrian recognition. But the speed is not fast due to the exhaustive search \cite{lan2022class}. Thus, this paper presents the optimization of the exhaustive search in HOG and SVM.  
    

\par The main requirements for our methods are that they work on the low computational hardware in real-time, and with high accuracy. To this end, we propose three methods including LBP features and Adaboost classifier, HOG features and SVM classifier, and Convolutional neural network for real-time pedestrian detection. In this section, we describe the details of the three methods and the experimental setup and dataset.

Choosing suitable hardware is a crucial and difficult task. After considering several alternatives, such as central processing units (CPU), graphic processing units (GPUs), digital signal processors (DSPs), and field programmable gate arrays (FPGAs) \cite{Kisacanin2008ECV}.
We have chosen the Intel NUC6i5SYK microcomputer as the hardware in this work since it offers a good trade-off between size, computing power, power dissipation, and price. In addition, Intel NUC6i5SYK is a general CPU and Linux (optional) platform with good compatibility that facilitates the portability of the algorithms to other hardware. It has physical dimensions of $111 \times 111 \times 35mm$, and embeds a 1.80 GHz Intel Core i5-6260U Processor. There are few applications where the NUC does not fit for mobile devices, in particular, drones, and small robots. In the experiments, we train the classifiers on the CPU computer and test the trained classifiers on Intel NUC6i5SYK. The experiments of HOG and SVM, LBP and Adaboost are implemented based on the OpenCV library. The Convolutional neural network is implemented based on Darknet.

\subsection{Dataset}
\label{subsec:data}

\par Data sets are a fundamental tool for comparing detection algorithms, fostering advances in the state-of-the-art. The INRIA pedestrian dataset is a traditional pedestrian dataset. The pedestrians in the image have relatively regular postures. Most of them are the persons walking or standing captured from different angles without much deviation from the horizontal perspective. The INRIA person data set \cite{Dalal2005Histograms} and  VOC (VOC2007 \cite{everingham2015pascal}, VOC2012\cite{everingham2015pascal}) are very popular in the Pedestrian recognition community, both for training detectors and reporting results. Compared to the INRIA dataset, VOC2007 and VOC2012 are more challenging. Although the VOC datasets originally provided samples and labels of 20 classes of objects, one of them is a person. we keep those labels for the class of person. 
The pedestrians in the images of VOC have a large variety of postures and are captured from different angles deviating severely from the horizontal perspective. To obtain a more comprehensive dataset, in this work, we combine the pedestrian data sets VOC(VOC2007, VOC2012) and INRIA as the data sets to train and test our methods. The statistics of the data sets are shown in table \ref{tab:dataset}.

\begin{table}[!ht]
\renewcommand\arraystretch{1.2} 
\caption{The statistics of the data sets. The training shows the number of images for the training data set. The testing shows the number of images for the testing data set. The $lab_{tr}$ shows the number of total labels on training images. The $lab_{te}$ shows the number of total labels on testing images.}
\label{tab:dataset}
\begin{tabular}{cccccc}
\hline
      & ~~training  & ~~$lab_{tr}$ & ~~testing  & ~~$lab_{te}$  & ~~total \\ \hline
INRIA & ~~614      & ~~1958       & ~~288      & ~~605         & ~~2563   \\
VOC   & ~~6095     & ~~13256      & ~~2007     & ~~4528        & ~~17784  \\
total & ~~6709     & ~~15214      & ~~2295     & ~~5133        & ~~20347  \\ \hline
\end{tabular}
\end{table}

As shown in table \ref{tab:dataset}, in this work we use 614 images from the INRIA dataset and 6095 images from the VOC dataset as the training data sets, 288 images from the INRIA dataset and 2007 images from the VOC dataset as the testing data sets.
For the training process, we use the bootstrap approach to train the SVM and AdaBoost classifiers until the classifiers perform well.  

\subsection{LBP features and AdaBoost classifier}
\label{subsec:ada}

\par Originally, the computation of the LBP feature is based on the comparisons of adjacent single pixels\cite{ojala1996comparative}. However, the strategy provided by OpenCV is based on the comparison between adjacent and small batches consisting of several adjacent pixels. Since the value of a batch is the sum of adjacent pixels in rectangle areas, it can be computed by using the integral image. This is also a new strategy which promotes the speed of the algorithm and it does not affect the accuracy of detection much.
\par The original strategy is shown in Fig.1. We generate the binary pattern based on the single pixel directly. Each pixel around the central pixel, whose value is 112, compares with the central pixel. If it is bigger, the binary value is 1. Otherwise, the binary value is 0.
\par The new strategy adopted by OpenCV is shown in Fig.2. Matrix2 is filled with the summations of every 4 adjacent pixels in matrix1. The binary pattern is calculated based on the comparisons of 8 non-central summations with the central summation.
\par The size of the image is \(S=height\times{width}\). The time complexity of generating the local binary pattern for the image is \(f(8S)\) in the original strategy. In the new strategy, the time complexity for calculating the integral image is \(f(S)\) and for the generation of LBP based on summations is 
\begin{equation}
	f(\frac{S}{4}\times{8})=f(2\times{S})
\end{equation}
Thus, the total complexity is \(f(3S)\). It is smaller than the time complexity of the original strategy.

\subsection{HOG features and SVM classifier}
\label{subsec:svm}


\begin{algorithm}[h!]
\SetAlgoLined
\DontPrintSemicolon
\SetKw{Define}{Define}
\SetKw{Select}{Select}
\SetKw{Extract}{Extract}
\SetKw{Init}{Initialise}
\KwData{5885 sample ROIs}
\Extract{ROI image features}
\Init{NEAT with parameters}
\Init{$P_0$ of ANN}
 \While{not $P=P_{\textrm{max}}$ or $f_g=f_{\tau}$ for some $g_i$}{
   \textbf{Select} 500 random sample ROIs\;
  \textbf{Evaluate} $g_i \in P_G$ \;
  \textbf{Select} sub-group $g \subset P = \{g_i, ...,g_j\}$ for reproduction based on $f_g$\;
 \textbf{Recombine} cross-over $g$ and create $g^*$\;
 \textbf{Mutate} $g^*$ with $p_{\textrm{mut}}$\;
  \textbf{Define} $P_{G+1}$\;
 }
 \Select{ $\max_{g_i \in P}f_{g}$}
 \BlankLine
 \caption{Evolving a real-time object recognition NN.}\label{al:NEAT}
\end{algorithm}


\par The initialization of cache and preprocessing for feature computation of HOG is based on the fact that the range of value of pixels is fixed, which is [0,255]. Thus, the gradients of \(dx\) and \(dy\) must be in the range of [-255,255].
\par The computation of the 9 bins of the Histogram of Gradient is based on gradients, which are \(dx\) in the x-direction and \(dy\) in the y-direction for each pair of adjacent pixels. The gradient for each point has magnitude and angle formed by \(dx\) and \(dy\). Then, for each gradient, there must be 2 bins adjacent to it. The magnitude votes the weights for the 2 bins based on linear interpolation. 
\begin{figure}
  \centering
  \includegraphics[width=0.5\textwidth]{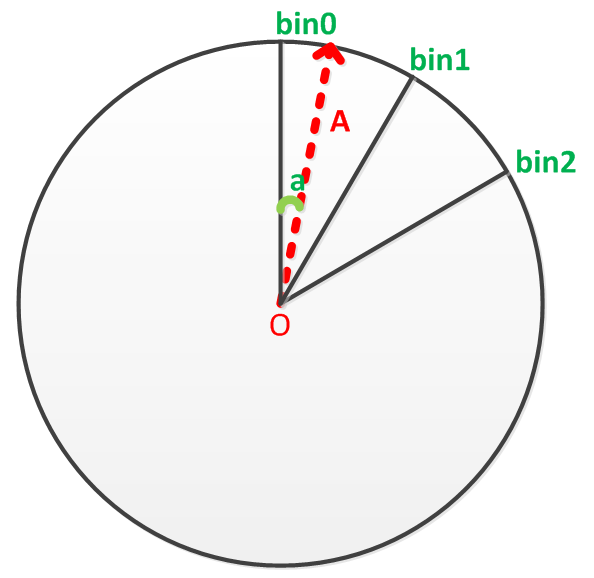}
  \caption{Gradients and adjacent bins in HOG}
\end{figure}
\par In Fig.3, O is a single point in an image and A is its magnitude. bin0 and bin1 are the 2 bins adjacent to this magnitude. Since we know the angle of the gradient, we can obtain the angle formed by bin0 and the direction of the gradient. Thus, the voted weights for the 2 bins can be computed. For each point, we need to store its 2 voted weights and the indices of the 2 bins.
\par As we mentioned above, given the \(dx\) and \(dy\) for a specific point, all the values above are fixed. And since the values of \(dx\) and \(dy\) are in the range of [-255, 255], all the data can be precomputed and stored in the hardware. Thus, these data can be pre-loaded before the computation of the HOG feature. This mechanism accelerates the process of feature computation up to 10\% faster than its original speed.
\par Except utilizing the cache data, we also adjust the parameters for the whole process of detection by HOG+SVM. We examine and test 3 parameters for the whole process, including the size of the step of scanning the whole image, the number of levels in the image pyramid, and the size of the samples for training and detection. We try different combinations of different parameters in order to balance the accuracy and efficiency of the program.

\subsection{Convolutional Neural Network}
\label{subsec:cnn}

\par For the experiments of HOG+SVM and LBP+Adaboost, we modify the dataset to improve the efficiency of the programs. We resize each sample provided by INRIA dataset from the size of ${96}\times{160}$ to ${32}\times{64}$ and resize the sample from VOC to ${32}\times{64}$ strictly despite its large variety of different sizes. Since HOG focuses on gradients and LBP focuses on the comparison of adjacent pixels, three channels of RGB have a subtle influence on the results. Thus, we transform all the RGB images into gray-scale images. This condition can also reduce the computation expense and improve the efficiency of the program.
\par The disadvantage of the models of LBP+Adaboost and HOG+SVM is that the exhaustive search 
does not consider the distortion of a sample with the original size which does not have the standard proportion of ${32}\times{64}$. In the experiment of HOG+SVM, we use the size of ${32}\times{64}$ as the standard size of samples. However, images can be distorted since we do not just consider samples with such size. We collect all the images with a person inside and resize them to ${32}\times{64}$ without considering the proportion.

\section{Experiments}
\label{sec:Res}

\subsection{The results}
\label{subsec:res}
\par Evaluating the quality performance consists of 2 values, including FPPI(false positive per image) and miss rate. The results of the detections for an image can only have 3 cases: detection matching the labelled areas(true positive); detection not matching any labelled areas(false positive); the labelled areas without any detection(miss positive). We focus on the last 2 cases which are both negative results. The number of detections not matching any labelled areas is the value of FPPI. The number of labelled areas without any detection is the value of the miss rate. The final goal of the pedestrian detection algorithm is to reduce them to become as low as possible. The lower they are, the better the results are.

For each image, we have labelled areas and detected areas and both of them are rectangles. 
\begin{itemize}
	\item To calculate FPPI, we need to count the detections that do not match any labelled areas in an image. When the overlap area in the labelled area and a detected area is less than 50\%, the detection does not match any labelled area, which is a false positive result\cite{everingham2010pascal}.
    \item To calculate the miss rate, we need to count the labelled areas without matching any detections in an image.
\end{itemize}

In order to describe the quality performance exclusively and the relationship between speed and quality performance specifically, we use 2 kinds of graphs to describe our experimental results.
In this plot, the x-axis represents the average time consumed by the model, the y-axis denotes the number of occurrences of matching between the labelled area and the detection area(Fig.4). It is said that when the intersection area is bigger than half of the labelled area, it is regarded as match\cite{everingham2010pascal}. This plot aims to present the relationship between speed and quality performance. The higher the value is, the better the performance is.


\begin{figure}
  \centering
  \includegraphics[width=0.25\textwidth]{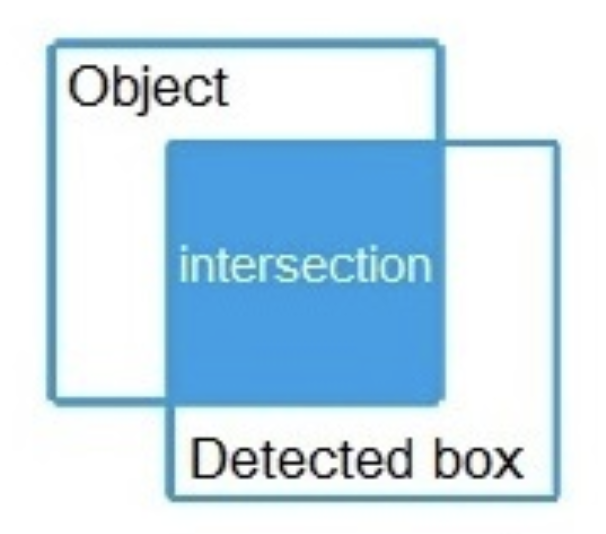}
  \caption{Definition of intersection area between labeled area and detection area}
\end{figure}



\par The efficiency of LBP+Adaboost is the highest among all of the three methods. Since the computation of the feature is the fastest and the process of running a decision tree is also fast. Moreover, according to the experiment, its accuracy of detection is close to HOG+SVM. 
We assume that is because we do not have enough training data to explore the further differences between them.
The model of Convolutional Neural Network based on darknet has the highest accuracy. 


\subsection{The Improvements}
\label{subsec:imp}

\par The experimental results demonstrate that our methods work well for real-time pedestrian recognition on the low computational CPU platform. However, we are still inspired by many approaches to improve the speed of pedestrian recognition. In the next work, we plan to implement the following approaches for faster speed on the smaller computers with lower computation that work on the devices with more constraints. 

For the model of HOG+SVM, we expect that the accuracy can be further improved by applying a Gaussian spatial window to the scale of the size of the sample, no application of a Gaussian spatial window to blocks. This is also a way to improve its efficiency.
The computation of HOG has considered too many factors, including Gaussian Window Application, magnitude and angle of gradients, and different sizes of blocks, cells and samples. Some of these features are redundancy. They need to be further experimented and extracted the most critical factors of affecting the result. 
For example, the experiment in C4\cite{wu2011real} has mentioned that the comparison of neighbour pixels is more important than the comparison of magnitude. Still, we do not know what kind of information extracted by HOG is the most critical to make it become such successful for human detection. If we can extract just one or two of the most significant information for feature computing, we can make a breakthrough in efficiency and accuracy. 
\par The strategy of computing LBP feature with the basic unit is the batch the pixels instead of a single pixel. It can also be applied to the computation of HOG in further research.
\par The accuracy of convolutional neural networks based on darknet is better than those of other models obviously. This is mainly attributed to the strategy of generating features for convolutional neural network models. The feature of convolutional combining with max pool is more general than other feature including HOG and LBP, which strictly relies on the shapes and outlines of the objects.

\section{Conclusions}
\label{sec:con}
Since the computational capacity of the CPU is limited, we can only promote efficiency by improving some processes in the program.
In this work, we investigate real-time pedestrian recognition on small physical-size computers with low computational resources for faster speed. This paper presents three methods that work on the small physical size CPUs system. First, we improved the Local Binary Pattern (LBP) features and Adaboost classifier. Second, we optimized the Histogram of Oriented Gradients (HOG) and Support Vector Machine. Third, We implemented fast Convolutional Neural Networks (CNNs). The results demonstrate that the three methods achieved real-time pedestrian recognition at an accuracy of more than 95\% and a speed of more than 5 fps on a small physical size computational platform with a 1.8 GHz Intel i5 CPU. Our methods can be easily applied to small mobile devices with high compatibility and generality.
In future, we will apply other AI technologies to improve this work such as knowledge graph \cite{lan2022semantic,liu2022towards}.

\bibliographystyle{ieeetr}
\bibliography{example}

\end{document}